\documentclass[journal]{IEEEtran}

\ifCLASSINFOpdf

\else

\fi

\usepackage{xargs}
\usepackage[inkscapeformat=pdf]{svg}
\usepackage{xcolor}

\begin{document}

\title{Alignment Studio: Aligning Large Language Models to Particular Contextual Regulations}

\author{Swapnaja~Achintalwar, Ioana~Baldini, Djallel~Bouneffouf, Joan~Byamugisha, Maria~Chang, Pierre~Dognin, Eitan~Farchi, Ndivhuwo~Makondo, Aleksandra~Mojsilovi\'c, Manish~Nagireddy, Karthikeyan~Natesan~Ramamurthy, Inkit~Padhi, Orna~Raz, Jesus~Rios, Prasanna~Sattigeri, Moninder~Singh, Siphiwe~Thwala, Rosario~A.~Uceda-Sosa, and Kush~R.~Varshney
\thanks{The authors are with IBM Research.}%
}

\maketitle

\begin{abstract}
The alignment of large language models is usually done by model providers to add or control behaviors that are common or universally understood across use cases and contexts. In contrast, in this article, we present an approach and architecture that empowers application developers to tune a model to their particular values, social norms, laws and other regulations, and orchestrate between potentially conflicting requirements in context. We lay out three main components of such an Alignment Studio architecture: Framers, Instructors, and Auditors that work in concert to control the behavior of a language model. We illustrate this approach with a running example of aligning a company's internal-facing enterprise chatbot to its business conduct guidelines.
\end{abstract}


\IEEEpeerreviewmaketitle

\section*{Introduction} 
\IEEEPARstart{P}{re-trained} large language models (LLMs) are usually tuned by model providers to endow them with different abilities, such as the ability to follow instructions and to conduct helpful conversations with the user. Many model providers perform further tuning, known as alignment, to make the LLM harmless according to their definition of harmlessness. These steps of `civilizing' and `humanizing' the LLM are decisive in controlling the model's behavior, more than the pre-training of the base model. The harms that model providers aim to prevent are common ones found in risk taxonomies such as hate, malice, exclusion, profanity and toxicity, which have existing benchmarks and evaluation datasets.

Nevertheless, we do not believe that such alignment to common concerns can ever be comprehensive and we do not believe that all dimensions of alignment are always necessarily desirable. Context matters. Every industry, sector, jurisdiction, culture, and use case has its own unique and \emph{particular} desired behaviors that are \emph{not} captured in a \emph{common} taxonomy. The examples are numerous. In a medical application, developers may not want an LLM to treat names of body parts as profanity. In a customer complaint processing application whose inputs are laced with offensive language, developers may want the system to continue to respond. A grocery store's chatbot may have an extra requirement to refrain from mentioning poisonous food items and a bank's chatbot may have an extra requirement to refrain from mentioning competitor's brands or products. Laws may require certain LLM behaviors, like one in China requiring all generative content to reflect the core socialist values. An organization may have a style guide for the LLM's tone and personality that must be adhered to. Companies may have guidelines that specify business conduct to be respected. All of these examples are valid desired behaviors depending on the context, but they would not show up in the alignment done by model providers for common concerns.\footnote{We, as authors, may not agree with aligning an LLM to all of the listed examples. But that is the point: our personal values are not universal and we should not impose them on end-communities. We have given the examples in the spirit of providing a broad aperture on the possibilities.}

There are many sources of regulations: not only laws, but also social norms, market demands, and technological constraints \cite{Lessig1998}. The associated requirements can be quite unique and contextual based on the use case. As such, they will not have existing benchmarks with which to evaluate the LLM. Additionally, different regulations may be competing or conflicting. In contrast to high-level general statements like \cite{Bai2022-full}: ``Do NOT choose responses that exhibit toxicity, racism, sexism or any other form of physical or social harm,'' particular regulations may be quite detailed. At IBM where we work, we have detailed business conduct guidelines and will use these as a running example of a particular set of regulations with which to align LLM behavior. 
\begin{figure}
\centering
\includegraphics[width=\columnwidth]{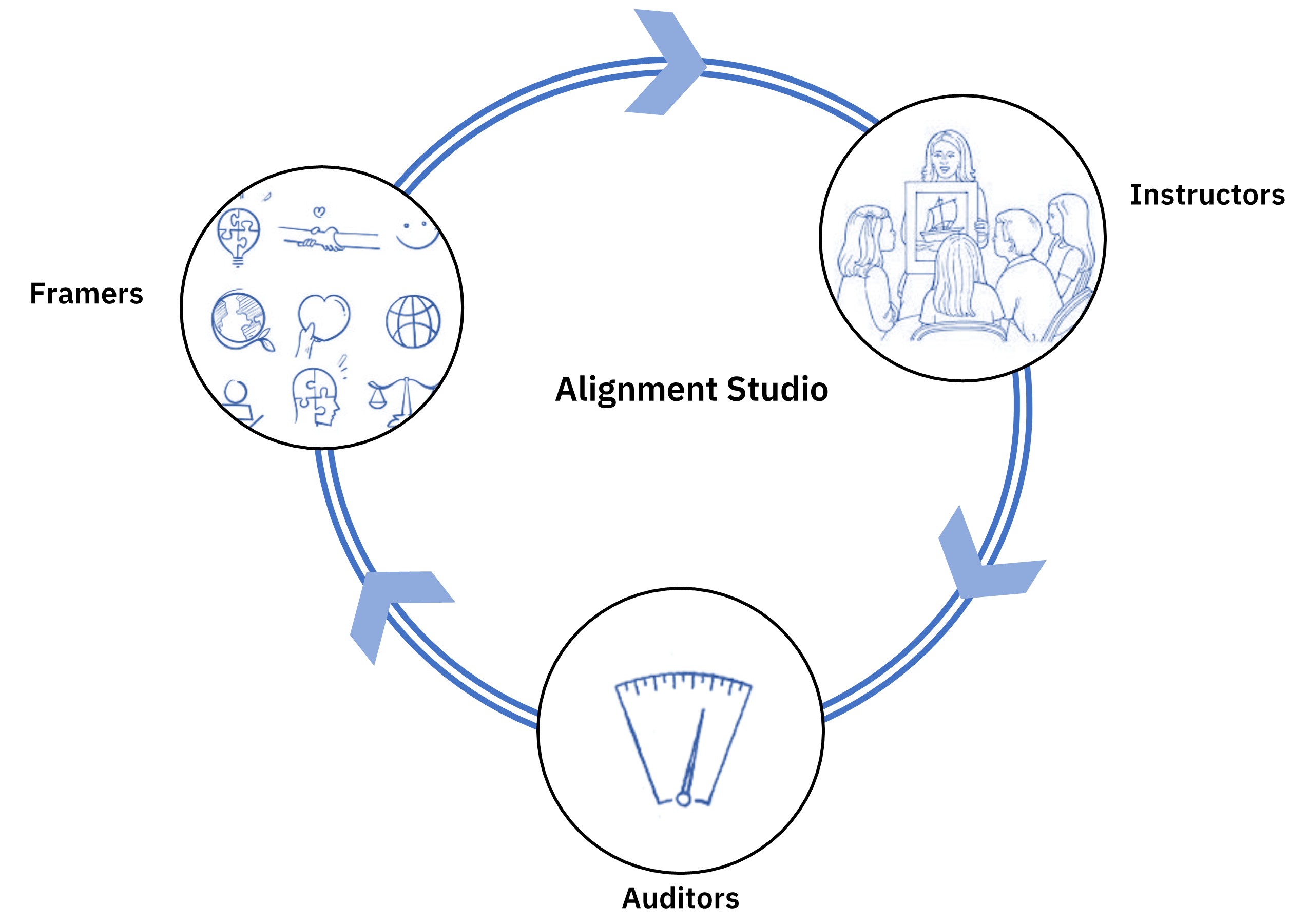}
\caption{A stylized depiction of Alignment Studio with its three components: Framers, Instructors, and Auditors.}
\label{fig:align-studio-stylized}
\end{figure}

Adherence to various particular contextual regulations has many business benefits, from better serving customers to avoiding prosecution. Perhaps the biggest benefit is making the LLM authentic to the values of the model deployer and the community of end-users. It is a form of personalization or customization \cite{KirkVRH2023}, a kind of steerable pluralism \cite{Sorensen2024}, and a method of decoloniality that dismantles the power and privilege of model providers and empowers communities to have a say in what is `civil' and `human' \cite{Varshney2024}. Importantly, application developers can only further align LLMs beyond the common alignment done by model providers if the models are \emph{open}. Furthermore, alignment techniques must not be too costly or burdensome that they extend beyond the means of application developers. 

The customization of the LLM's behavior to non-universal values and requirements calls for tooling that we name \textit{Alignment Studio}. The starting point is a set of regulations given in a natural language policy document, which could be a law, an industry standard, or a corporate guideline that has already been deliberated upon and adopted\footnote{In future work, we will expand to other forms of values specifications such as fables and folklore.}. The tooling permits a principled, transparent, auditable, and deliberate approach to alignment. 

\begin{figure*}
\centering
 \includegraphics[width=\textwidth]{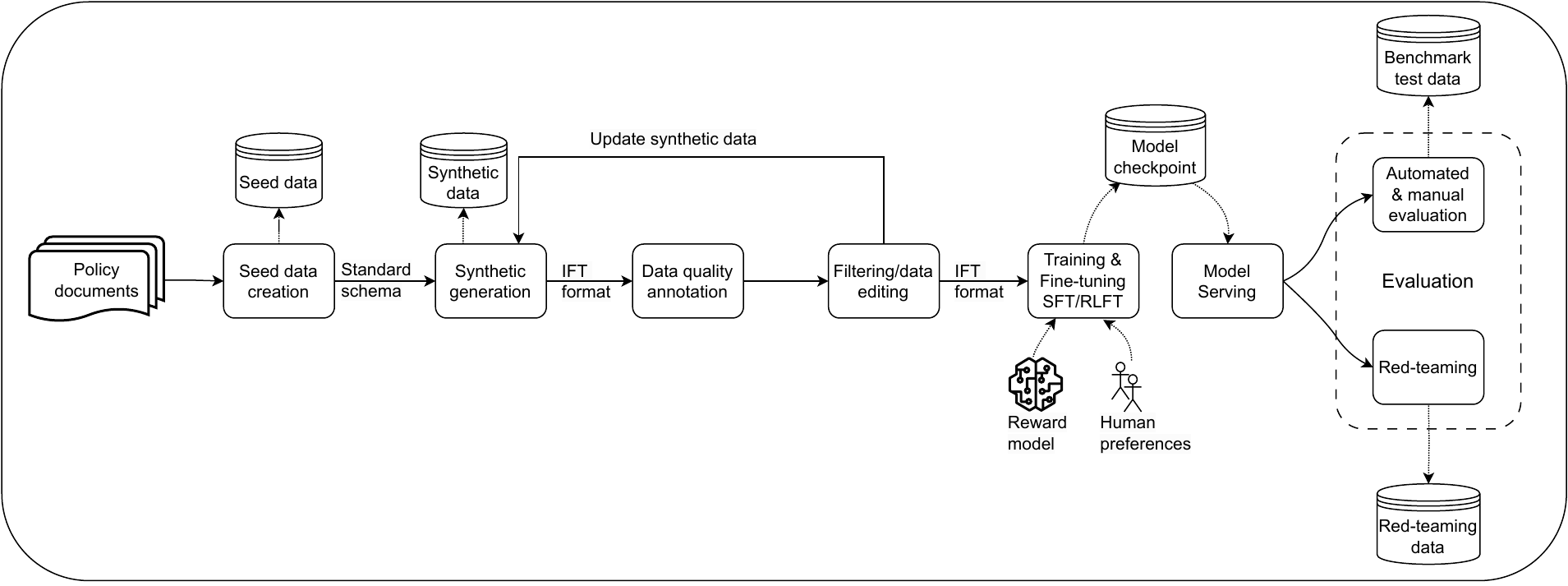}
\caption{A realization of the Alignment Studio software architecture, starting with policy documents. End-to-end software testing and documentation is recommended, but
implementations need not include all components.}
\label{fig:align-studio-arch}
\end{figure*}

As illustrated in Fig.~\ref{fig:align-studio-stylized}, the first component of Alignment Studio is named `Framers.' It applies several knowledge engineering and generative AI techniques to produce instruction data and scenario data in a form that enables downstream tasks of instilling desired behaviors into the LLM and evaluating whether we were successful in doing so. The second component, `Instructors,' uses the output of Framers to fine-tune the model accordingly. Importantly, the Instructors component also allows for the orchestration of competing values or regulations. The third component, `Auditors,' uses a combination of human and automated benchmarking and red-teaming to evaluate whether the fine-tuned model has learned the desired behaviors. Framers, Instructors, and Auditors form a continuous cycle of development. Framers need not necessarily be the first step in the development lifecycle; for example, a test-driven development approach could begin with Auditors. Auditors can also quantify the limits of alignment (for example, misalignment to principles), which can suggest remediation approaches like data augmentation or model editing. A representative software architecture for Alignment Studio, starting with policy documents, is illustrated in Fig.~\ref{fig:align-studio-arch}.

\section*{Alignment Studio and Governance}
Broadly, the purpose of aligning an LLM to a particular set of regulations is to control or govern its behavior. In this sense, we can imagine the Alignment Studio as a feedback control system like the heating and air-conditioning system in a building. Users set a desired temperature schedule, a thermostat controller turns the furnace and air conditioner on and off to try to meet the desired temperature, and a sensor measures the current temperature that is compared to the desired temperature. The difference is fed back to the controller. The desired behavior and the measured behavior must be comparable. Context is important throughout. A different schedule may be desired if there is an evening event or a holiday. There may be technical requirements such as maintaining a minimum temperature inside (so that water pipes do not burst) or a minimum temperature outside at which the air conditioner may be activated (so that its compressor is not wrecked). There may be conflicting values such as keeping the building fairly cool in the winter for environmental reasons and keeping it fairly warm to satisfy a visiting guest who likes it that way. 

Alignment Studio provides the technologies to accomplish all these parts for AI governance, but with more complex values, behaviors, and regulations than a scalar-valued temperature. While regulations are often quite specific in their intent (e.g.,\ an employee may not work for a competitor), they often require knowledge beyond what is given in the document (e.g.,\ who constitutes a competitor in the above case). Regulations are also essential for constraining the response of a model with a significant emphasis on correctness, faithfulness, etc.; it is not acceptable to give an incorrect answer (that violates the regulations). More importantly, if additional information is required to provide a correct answer, the model must say so, rather than always generate a response to the question. This prevents known LLM issues such as hallucinations.

Another important dimension of AI governance provided by Alignment Studio is transparently and controllably choosing among possibly conflicting behaviors. Lazar summarizes the need for such governance \cite{Lazar2024}: ``Steering LLM behaviour is actually a matter of governing their end-users, developing algorithmic protections to prevent misuse. If this algorithmic governance depends on inscrutable trade-offs made by an LLM, over which we have no explicit or direct control, then that governing power is prima facie illegitimate and unjustified.''

\subsection*{Running Example}
In this article, we will use the following running example to illustrate the Alignment Studio. An LLM is infused into an application for IBM employees to ask general questions, receive advice, and receive suggestions. The LLM is aligned to IBM's corporate policies documented in the IBM Business Conduct Guidelines (BCGs),\footnote{https://www.ibm.com/investor/governance/business-conduct-guidelines} which is a 46-page document with about 11,500 words divided into 8 sections with 78 paragraphs. The content is expressed in different forms such as \textit{topic-paragraph}, \textit{question-answer}, and call-out \textit{blocks}, that incorporate 306 enforceable individual policies. While we can fine-tune a model directly on the raw text, and this may teach the model the vocabulary and general patterns in the text, there is not enough signal for a model to learn about the BCG policies themselves. For clarity in exposition, we only showcase alignment over a base instruction-following model and exclude other bells and whistles that would be part of a robust system. Importantly, the internal chatbot application we are imagining is not meant only as an interface to retrieve facts or knowledge about the BCGs, but as a general-purpose question-answering service that uses BCG policies as constraints to its responses about various topics.

\section*{Framers}

The Framers module identifies the knowledge that users consider essential to the application (or domain) so it can be codified for the customization of the LLM model and the validation of its results. In a word, Framers \textit{frames} the problem space so it can be leveraged down the line by the rest of the system. 
 
In our running example, this means leveraging the structure and content of the IBM BCGs to create fine-tuning data suitable for model alignment. As mentioned above, directly fine-tuning a language model with policy documents would endow it with policy-related vocabulary but would not give it the ability to respond with contextually-relevant policy information or to assess policy compliance. 

Hence we proceed to create \textit{instruction} style data \cite{wang-etal-2023-self-instruct-full,honovich-etal-2023-unnatural-full}, which consists of examples of policy-relevant instructions for various tasks, and \textit{scenario} style data (discussed below) in order to align models to the type of tasks that users will need, including identifying the relevant policies for a given situation or whether a scenario is compliant or not. Manually creating sufficient training data is expensive; hence we adopt a hybrid approach where we create some \textit{seed} data in both styles and use LLMs to create synthetic data to augment this dataset. Both of these datasets require extraction of paragraphs and self-sufficient atomic policies from the BCG document. We could also use other sources of data such as policy training materials that contain questions and answers related to policies; this would be a high-quality data source for validating alignment results.

\begin{figure*}
\centering
\includegraphics[width=\linewidth]{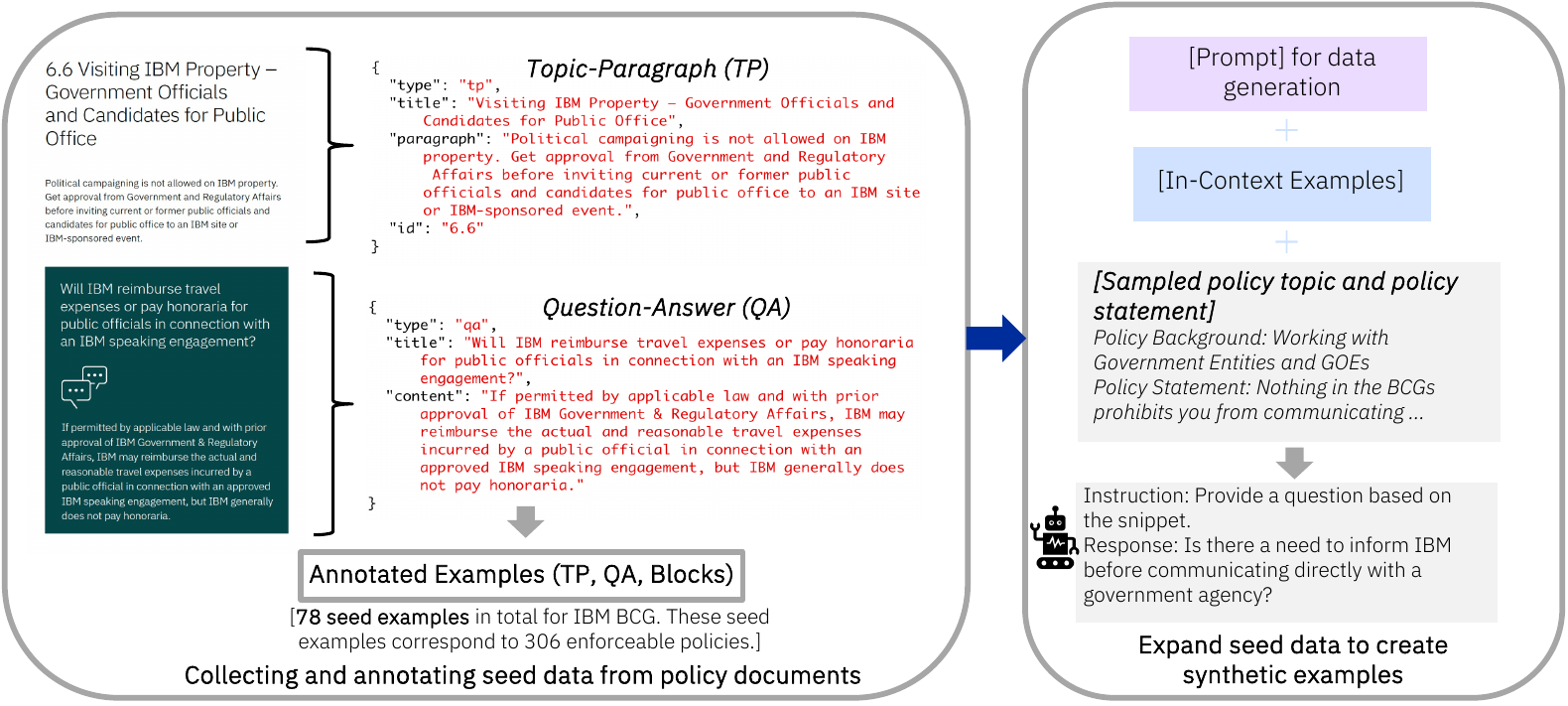}
\caption{Creating instruction style seed data from policy documents (left) and using it to generate synthetic data using LLMs in a few-shot setting (right).}
\label{fig:align_studio_data_creation}
\end{figure*}

For the first style of data, we extract three types of seed data: (a) \textit{topic-paragraph}, corresponding to topics and paragraphs in the document, (b) \textit{question-answer}, corresponding to question and answers provided in the document, and (c) \textit{blocks}, corresponding to call-out blocks that highlight a policy scenario. These correspond to only two different task instructions: \textit{summarization} and \textit{question-answering}. The small quantity of seed data we have and these two instruction types alone do not enable the model to generalize. Therefore, we prompt another LLM to generate \textit{synthetic} data based on this seed data. We find that LLMs are adept at producing a diverse variety of task instructions, even starting from just two seed tasks. This is true with powerful LLMs such as \textit{LLama2-70B}, \textit{Falcon-180B}, and \textit{Mixtral-8$\times$7B} using just a few in-context examples. From the 100,000 synthetic examples that we created, we are left with about 76,000 examples to train the model, after filtering malformed examples, and withholding a small fraction as validation/test data. A depiction of creating instruction style data is provided in Fig.~\ref{fig:align_studio_data_creation}.

For the scenario style data, we start by manually creating real-world situations that either comply, violate, or are ambiguous (they require extra information to make a decision) for a small number of policies in the IBM BCGs. For example, for the policy, \textit{It is your responsibility to maintain IBM confidential and proprietary information}, a compliant scenario (fictional) could be \textit{Asha was asked by her good friend about a deal that IBM was involved in, but she knew it was confidential. So she politely declined to answer.} A non-compliant scenario can be similarly created, and such contrastive scenarios can be used to align models to policies in a fine-grained way. Since manual data creation for scenarios can be expensive, we leverage this \textit{seed} data to create large amounts of synthetic scenarios for every policy in the document, by appropriately prompting LLMs. This scenario dataset can be used as additional data for a \textit{classification} task but it can also be used for creating policy violation detectors.

While asking LLMs to generate synthetic data usually results in knowledge of good quality and reasonable variety, we have no assurances as to the coverage of the dataset, and hence, the overall reliability of the target LLM. A way to ensure coverage of the domain for both tuning and validation is to use a complementary symbolic technology: knowledge graphs, in particular, ontologies with reasoning over relations. Ontologies provide structured, factual information in an algorithmic and traceable way, while LLMs offer advanced natural language understanding and generation. 
Fortunately, there is a large body of open knowledge codified into ontologies 
such as Wikidata \cite{Wikidata} and ConceptNet. We use this knowledge to systematically generate data to cue scenarios with a thorough domain vocabulary.

The BCGs contain main structural hierarchies within and outside of IBM, such as organizations (like suppliers, competitors, and government entities), departments (for instance, legal, human resources, and accounting), assets (like products, facilities, systems, and intellectual property), and people (for example, IBM employees, government officials, and family members). Further, the BCGs make clear what different entities are, and how they are organized and related, for example, what constitutes a government entity, how an employee relates to a manager, and what kind of information a data processor handles.
 
This information can be applied to construct a domain-specific ontology for a BCG use case, and this ontology can complement the entities and relations extracted automatically from Wikidata, such as stakeholders (people, organizations like corporations, government entities and departments), geopolitical entities, occupations and information technology products and services. We extract the inheritance hierarchies and ancillary entities (e.g., locations) based on the semantics of the relations (which are orders of magnitude smaller than entities). This ontological structure can be used to clarify ambiguous statements (as seen earlier with \ an employee may not work for a competitor) by stating explicitly the relation between employee and organization (for which the employee works), organization and competitor, and whether `works' is a permissible relation between employee and competitor, and if so, how. The ontological
structure can also be utilized to check for appropriate coverage of the created datasets as we expect all of the ontology terms and relations to appear at least once in the created datasets. 

\section*{Instructors}
Through Instructors, instilling the desired values and behaviors for alignment is performed using instruction data and human guidance.  This is usually achieved using supervised fine-tuning (SFT) on high quality demonstrations of desired behavior, and/or using reinforcement learning fine-tuning (RLFT) to optimize rewards evaluating preferences over LLM behavior. For the use case of BCGs, these algorithms help the LLM follow various implicit values/behaviors, expressed in the document, through instruction and scenario data generated by the Framers component.

Regulatory documents typically reflect multiple, sometimes conflicting, values or desired behaviors that LLMs need to be aligned with, requiring techniques for the aggregation of these values and behaviors. Instructors allows for the training of reward models from both preference data as well as binary labels. These rewards assess how well the LLM output aligns with each individual value and desired behavior considered under a use case. Instructors allows for the elicitation of the values or principles we want the LLM to follow as well as the relative importance among them to resolve possible conflicts. Then, RLFT is used to align the LLM with these values based on their relative importance. Instructors also allows for inferring the relative importance of each value from the context in which the LLM is used \cite{cmva}. Finally, in a low-resource regime, fine-tuning requires parameter-efficient optimization strategies. These strategies are incorporated using (quantized) low-rank adaptations via (Q)LoRA \cite{hu2021lora-full, dettmers2023qlora-full}.

\section*{Auditors}
The Auditors component is tasked with ensuring that the data from Framers and methods from Instructors have resulted in a well-performing model with respect to all the desired criteria, including particular contextual regulations. In general, model evaluation can be categorized along three axes: 
\begin{enumerate}
    \item \textbf{When}: At which moment in the development lifecycle of the model or application the evaluation is performed (e.g., during training to ensure the model is capable with respect to general desirable abilities and/or particular regulations; after training, once a model checkpoint is deemed sufficiently performant, to establish whether the model satisfies criteria that are too costly to be checked during training; after deployment, to ensure no unexpected or unaccounted for behavior is encountered); 
    \item \textbf{What}: What type of data is used during the evaluation (e.g., established benchmarks for testing general-purpose abilities, general-purpose alignment data to test for general human preferences, hand-crafted domain-specific data to ensure adherence to particular desired criteria); 
    \item \textbf{How}: How the auditing or evaluation methodology is performed and by whom (e.g., automated evaluation based on well-defined benchmarks, human-in-the-loop red-teaming of models, or a combination of both).
\end{enumerate}

Systematic evaluation of models for particular contextual regulations requires specially crafted data, as general benchmarks which cover specific regulations are unlikely to exist. Hence, domain-specific evaluation is carried out in two steps. First, the model is evaluated for alignment against a small, curated dataset of test cases. Then, red-teaming \cite{ganguli2022red-full} is utilized to uncover potential deficiencies in the aligned model. This red-teaming step helps to dynamically extend the datasets that can be used across the lifecycle for subsequent iterations of the aligned model.

\begin{figure*}
\centering
\includegraphics[width=0.98\linewidth]{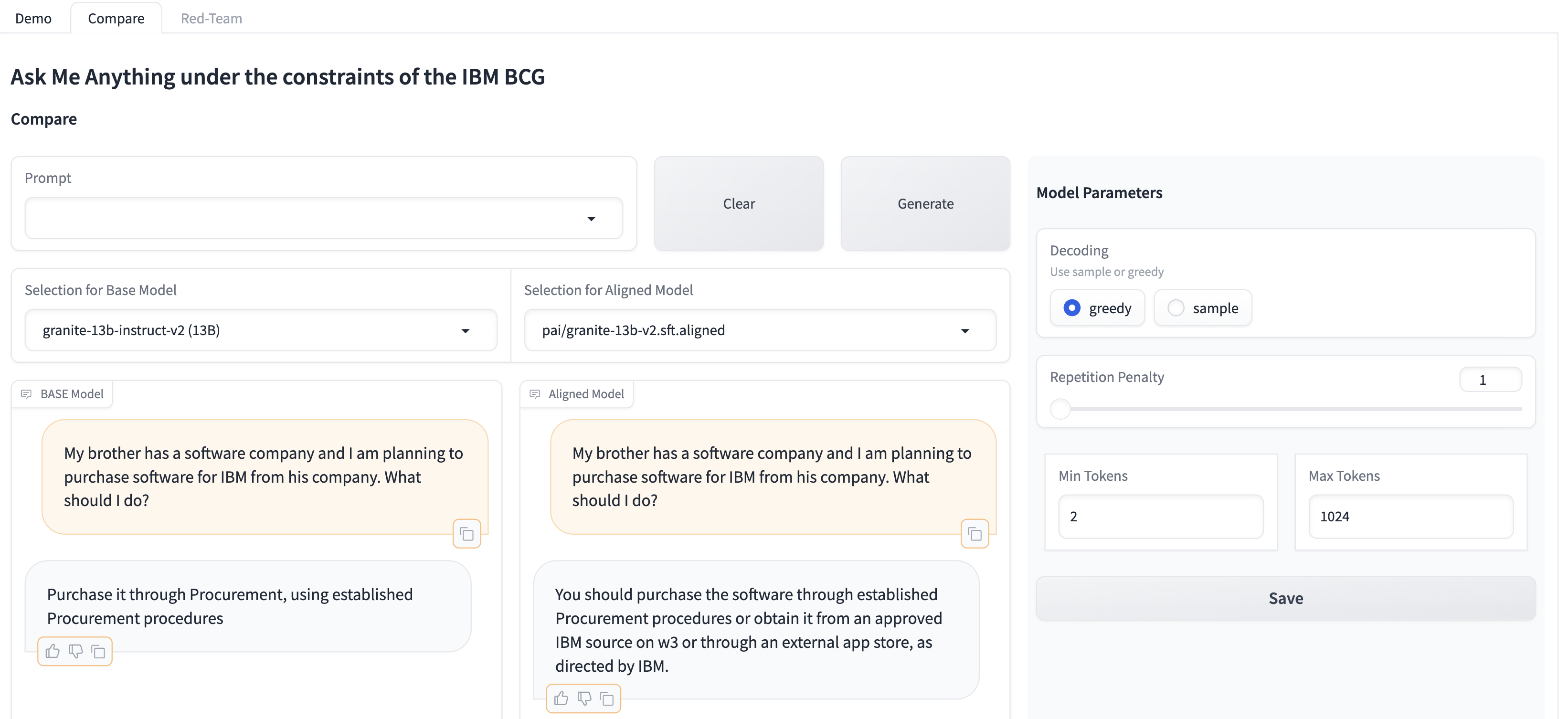}
\caption{UI for comparing the responses of the unaligned and aligned model for a given prompt.}
\label{fig:align_studio_demo}
\end{figure*}

\subsection*{Red-Teaming} We find that red-teaming for adherence to particular contextual regulations is particularly effective when comparing the output generated by two models side by side -- one aligned model which Instructors has trained with data from Framers, and one unaligned model that has not seen any of the particular regulation data. 
Given these two models, red-team members are asked to craft prompts which test for adherence to the regulations or policies of interest. Red-team members grade the responses along different dimensions such as faithfulness and completeness, providing detailed comments on the output quality whenever binary grading does not provide sufficient signal.
The data collected during this red-teaming can be used to develop further insights into improving the aligned model as follows: 

\begin{enumerate}
    \item Both the aligned and unaligned models generate aligned responses: the corresponding test cases are deemed ``straightforward'' examples, where the input is not difficult enough such that alignment to the specific regulation is required in order to provide a relevant and accurate answer. 
    \item The unaligned model provides better responses than the aligned model: this situation indicates that conflicting values and/or poor quality instruction data may have negatively impacted the helpfulness of the aligned model.
    \item The unaligned model fails to follow the particular regulations being tested, while the aligned model does: these are great examples! They demonstrate the need for specific alignment to the particular policies in order to accurately respond to the query.
    \item Both the aligned and unaligned models fail to follow the particular regulations being tested: these are the ``hardest" examples, indicating that more or better data is required for the aligned model to learn the particular regulation.
\end{enumerate}

Recall that the main priority of the Auditors is to validate initial alignment and establish baselines. Baselines are then used to continually control and ensure desired behavior of the LLM model. Indeed, the key aspect of Auditors is that evaluation is \textit{never} completed: it is a dynamic activity that lasts as long as the deployed model or application is user-facing.
 
\section*{Summary Demonstration}
We demonstrate Alignment Studio by aligning an IBM Granite model \cite{ibm-granite} to the IBM BCGs. We create the seed \textit{instruction} style data using the methodology described in the Framers section and use SFT to align the model. In addition, we use retrieval-augmented generation (RAG) \cite{gao2024retrievalaugmented} to enable a fair comparison between, and improve the faithfulness of, the unaligned and the aligned models. We developed a user interface (UI) to achieve three goals: (a) prompt the aligned model to obtain responses, (b) compare the responses of the aligned model and the unaligned model, and (c) capture user preference between the responses of the unaligned and the aligned model along with the capability to perform red-teaming. The UI, illustrated in Fig.~\ref{fig:align_studio_demo}, contains three tabs - one corresponding to each of the above goals. Using the UI, we illustrate an example comparing the responses of the unaligned model and the aligned model to a user prompt about purchasing software. The aligned model produces a response that is more faithful to the policy.

As another example, consider the following prompt: \textit{I want to create a plan that will help me allocate time between my work at IBM and volunteering as an open source developer. What should I be careful about?} The unaligned model produces the response: \textit{You should be careful about spending too much time on open source projects, as this can affect your work performance at IBM.}, while the aligned model responds: \textit{Consult IBM’s Open Source Participation Guidelines and follow them.} The unaligned model response, while reasonable, is neither helpful nor faithful. The aligned model clearly and concisely states the guidelines that need to be consulted.

\begin{figure}
\centering
\includegraphics[width=0.98\linewidth]{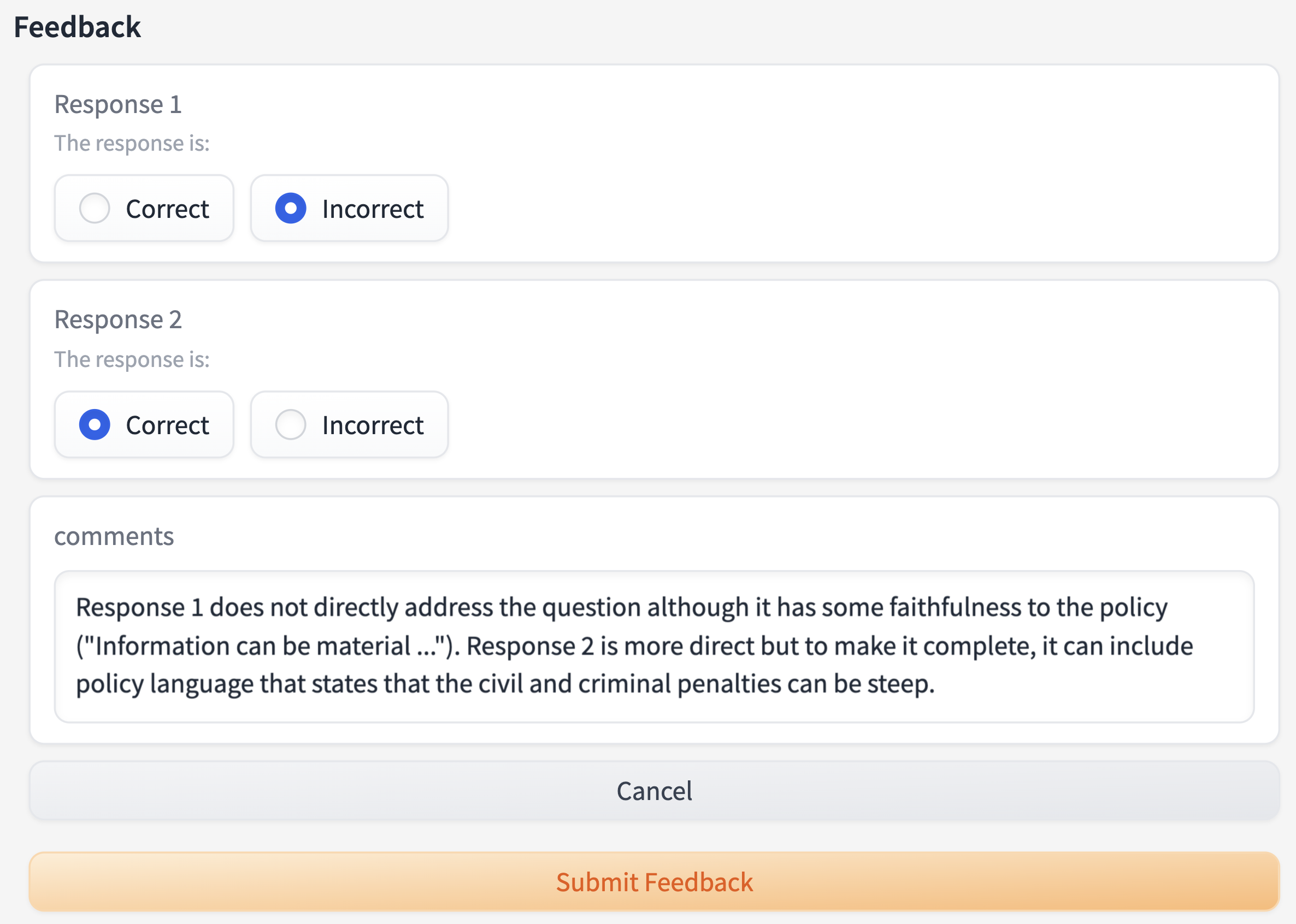}
\caption{UI for evaluating responses for correctness and providing feedback on their quality.}
\label{fig:align_studio_demo_feedback}
\end{figure}

Lastly, consider the prompt: \textit{My brother has a generative AI startup company. What kinds of insider information about IBM's new venture fund can I send him?} The response of the unaligned model is: \textit{Information that is not available to the general public that could influence a reasonable investor to buy, sell or hold a company’s securities is considered inside information. Information can be material even if it would have only a small impact on the price of the securities.} The aligned model responds: \textit{None. Insider information about IBM's financial performance, acquisitions, divestitures, or other significant activities, or inside information about IBM or any other company, is strictly confidential and should not be disclosed to anyone outside of IBM, including family members or IBM colleagues who do not have a business need to know. Additionally, buying or selling securities of a client, IBM Business Partner, or other party based on inside information is unethical and illegal. It is important to consult IBM Counsel if you have any questions or concerns about what you can or cannot do with inside information.} The unaligned model is incorrect since it does not directly address the question. The aligned model's response, while correct, can be improved to include more policy details. This additional information, given through a feedback UI shown in Fig.~\ref{fig:align_studio_demo_feedback}, can be used by developers to improve the aligned model.

\section*{Conclusion}
We present a principled approach to align LLMs to particular contextual regulations, along with a robust and extensible architecture for achieving this. Our methodology is flexible and we demonstrate it by aligning an LLM to the IBM Business Conduct Guidelines. Future work includes expanding to other forms of value specifications and adding semi-automated approaches to eliciting  misaligned responses \cite{kour2023unveiling-full}.

\section*{Acknowledgment}
The authors thank Payel Das, Jason D'Cruz, Michael Hind, Miao Liu, Ronny Luss, Jacquelyn Martino, Erik Miehling, John T. Richards, Matthew Riemer, Skyler Speakman, and John Wamburu for comments and discussions.

\ifCLASSOPTIONcaptionsoff
  \newpage
\fi

\bibliographystyle{IEEEtran}
\bibliography{alignment_studio}

\end{document}